\documentclass[10pt,twocolumn,letterpaper]{article}

\usepackage[camera-ready]{cvpr}      
\usepackage{multirow} 
\usepackage{xcolor}      

\usepackage{colortbl}    
\usepackage{arydshln}    
\usepackage[table]{xcolor}  
\usepackage{arydshln}    
\usepackage{booktabs}   
\usepackage{array}      

\usepackage{enumitem}

\usepackage{minted}
\usepackage[table]{xcolor}

\definecolor{BestBG}{HTML}{F4B8B8}    
\definecolor{SecondBG}{HTML}{EDE7F6}  
\definecolor{MyRed}{HTML}{F4B8B8}    
\definecolor{MyBlue}{HTML}{EDE7F6}   

\newcommand{\best}[1]{\cellcolor{BestBG}\textbf{#1}}      
\newcommand{\secondbest}[1]{\cellcolor{SecondBG}\underline{#1}} 
\definecolor{cvprblue}{rgb}{0.21,0.49,0.74}
\usepackage[pagebackref,breaklinks,colorlinks,allcolors=cvprblue]{hyperref}

\def\confName{CVPR}
\def\confYear{2026}

\title{Bridging Human Evaluation to Infrared and Visible Image Fusion}

\author{
  Jinyuan Liu\textsuperscript{\rm 1}, \quad Xingyuan Li\textsuperscript{\rm 2}, \quad Qingyun Mei\textsuperscript{\rm 3}, \quad Haoyuan Xu\textsuperscript{\rm 3}, \\  Zhiying Jiang\textsuperscript{\rm 4}, \quad Long Ma\textsuperscript{\rm 3}, \quad Risheng Liu\textsuperscript{\rm 3}, \quad Xin Fan\textsuperscript{\rm 3}\thanks{Corresponding author.}\\
  {\small\textsuperscript{1} School of Mechanical Engineering, Dalian University of Technology }\\
  {\small\textsuperscript{2} College of Computer Science and Technology, Zhejiang University }\\
  {\small\textsuperscript{3} School of Software Technology \& DUT-RU International School of ISE, Dalian University of Technology }\\
  {\small\textsuperscript{4} College of Information Science and Technology, Dalian Maritime University }\\
  {\tt\small atlantis918@hotmail.com} \hspace{0.1cm}
  {\tt\small xin.fan@dlut.edu.cn} \hspace{0.1cm}
}

\begin{document}
\maketitle
\begin{abstract}
Infrared and visible image fusion (IVIF) integrates complementary modalities to enhance scene perception. Current methods predominantly focus on optimizing handcrafted losses and objective metrics, often resulting in fusion outcomes that do not align with human visual preferences. This challenge is further exacerbated by the ill-posed nature of IVIF, which severely limits its effectiveness in human perceptual environments such as security surveillance and driver assistance systems. To address these limitations, we propose a feedback reinforcement framework that bridges human evaluation to infrared and visible image fusion. To address the lack of human-centric evaluation metrics and data, we introduce the first large-scale human feedback dataset for IVIF, containing multidimensional subjective scores and artifact annotations, and enriched by a fine-tuned large language model with expert review. Based on this dataset, we design a domain-specific reward function and train a reward model to quantify perceptual quality. Guided by this reward, we fine-tune the fusion network through Group Relative Policy Optimization, achieving state-of-the-art performance that better aligns fused images with human aesthetics. Code is available at \url{https://github.com/ALKA-Wind/EVAFusion}.
\end{abstract}    
\section{Introduction}
\label{sec:intro}


\noindent Infrared and visible image fusion (IVIF) stands as a pivotal technique in computational imaging, aiming to synthesize a single image that comprehensively preserves the thermal radiation information from infrared images and the rich textual details from visible images~\cite{zhang2025ddbfusion,yang2025instruction,zou2026contourlet,qin2026disentangle}. This complementary fusion paradigm is crucial for a wide spectrum of high-stakes applications, including autonomous driving~\cite{liu2023multi}, security surveillance~\cite{yi2024text}, medical imaging~\cite{zhao2023cddfuse}, remote sensing~\cite{yu2023difficulty}, and military reconnaissance~\cite{fang2025integrating}.

\begin{figure*}[t]
    \centering
    \includegraphics[width=1.0\linewidth]{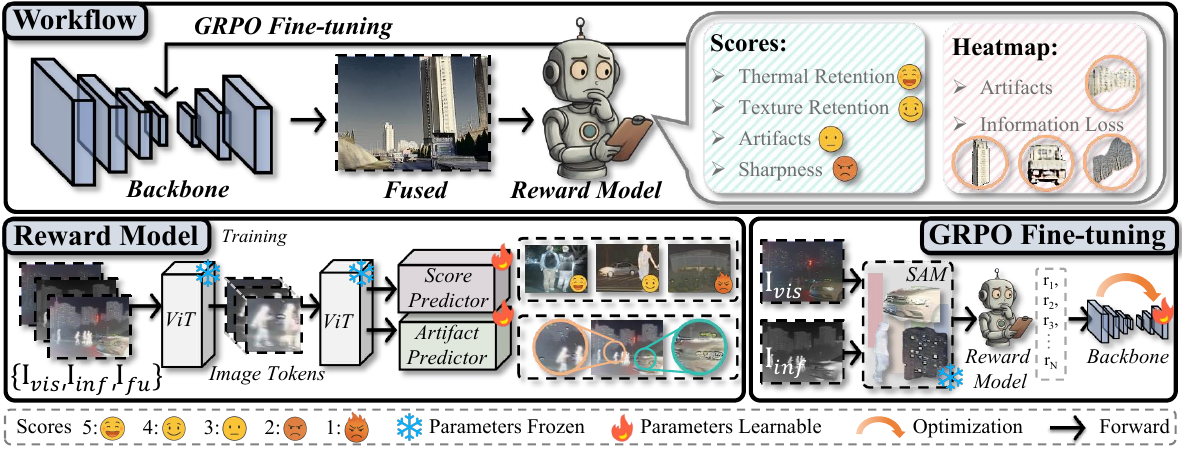}
    \caption{\textbf{Illustration of our pipeline.} The fine-tuning module is based on RLHF, integrating the ViT-based fusion-oriented reward model trained with the Human Feedback Dataset. A segmentation-assisted mechanism, based on GRPO, is introduced for fine-tuning, aiming to improve the quality of the fused images.}
    \label{fig:pipeline}
    \vspace{-3mm}
    
\end{figure*}

Despite significant advancements, the fundamental goal of IVIF is to generate a fused image that is preferable for human observation and analysis, remaining inadequately addressed. The task is inherently ill-posed due to the absence of a unique ground-truth fusion result~\cite{xu2020u2fusion}. Consequently, the field has long been dominated by the pursuit of optimizing handcrafted objective functions and numerical metrics, such as entropy, structural similarity, and gradient-based measures~\cite{xu2019learning,liu2022target,liu2024promptfusion}. While these approaches have driven progress, they suffer from a critical limitation: \textbf{the discrepancy between these mathematical proxies and genuine human perceptual preferences.}

Recent deep learning-based methods, particularly those employing generative models~\cite{ma2019fusiongan,li2024contourlet,qin2025robust,quan2024siamese,ma2024followpose} and attention mechanisms~\cite{li2025difiisr,li2023lrrnet}, have demonstrated powerful capabilities in feature extraction and fusion. However, they largely inherit the same evaluation paradigm, training networks with pixel-level or feature-level losses that are ultimately divorced from human judgment. This creates a significant gap: current fusion models are not explicitly optimized to produce outputs that align with human aesthetic and perceptual consistency. The root causes of this gap are twofold: (1) the lack of a large-scale, high-quality dataset containing human feedback on IVIF results, (2) the absence of a reliable and automated reward mechanism to quantify perceptual quality and guide the model's learning process.

To bridge this critical gap, we introduce a novel Feedback Reinforcement Framework that directly integrates subjective human evaluation into the IVIF pipeline. Our work is built upon the core insight that human preference should be the ultimate supervisor for this ill-posed task. The key challenge lies in efficiently and scalably incorporating this expensive, subjective feedback into the optimization loop.

Specifically, we first constructed a comprehensive human feedback dataset comprising subjective assessments and artifact annotations for fused images. This dataset includes 850 image pairs drawn from eight diverse datasets, fused by eleven representative models to generate 9,350 images. 
To efficiently and accurately annotate this large dataset, we employed GPT-4o based on expert-labeled samples.
Then, we leveraged the model to assist in scoring and annotation, with expert review to ensure quality. Using this dataset, we designed a specialized reward function and trained a reward model to capture human perceptual preferences quantitatively. Finally, guided by this reward model, we fine-tuned existing fusion networks using Group Relative Policy Optimization (GRPO), 
achieving state-of-the-art performance that better reflects human aesthetic judgments. Our main contributions are summarized as follows:
\begin{itemize}
\item[$\bullet$] We propose a feedback reinforcement IVIF framework that explicitly integrates subjective human preferences into the fusion process, effectively bridging the gap between objective image quality metrics and human perceptual consistency.

\item[$\bullet$]
To address the lack of human-centric evaluation data,
we construct the first large-scale, high-quality human feedback dataset in the IVIF domain, featuring multidimensional subjective scores and detailed artifact annotations.

\item[$\bullet$] 
We develop a reward function and a tailored reinforcement learning strategy for IVIF, enabling fusion models to better capture human visual preferences and achieve state-of-the-art performance.
\end{itemize}

\vspace{-5pt}
\section{Related works}
\label{sec:formatting}
\subsection{Infrared and Visiable Image fusion} 

\begin{figure*}[t]
    \centering
    \includegraphics[width=1.0\linewidth]{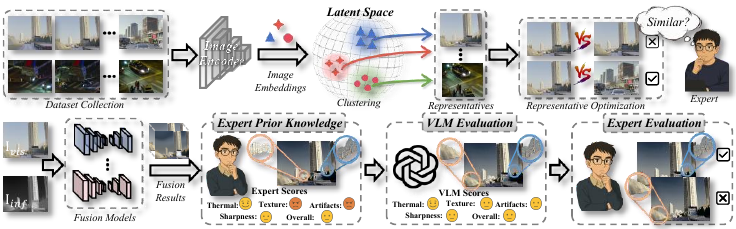}
    \caption{\textbf{Workflow of the dataset collection.} The dataset is processed and clustered to optimize representative images, followed by expert screening. Use fusion models to fuse $I_{vis}$ and $I_{inf}$ images, with experts scoring and annotating a subset as prior knowledge for aligning the GPT model. After GPT annotates all images, experts review the results, resulting in the human feedback IVIF dataset.}
    \label{fig:collect}
    \vspace{-3mm}
\end{figure*}


\noindent As an important image enhancement technique, infrared and visible image fusion has continued to attract increasing attention in recent years ~\cite{li2023text,cao2024conditional,li2025mulfs,li2024deep,cheng2025fusionbooster,sun2022detfusion,tang2023rethinking}. 
With the rapid development of deep learning, deep learning-based image fusion methods have gradually become a research hotspot. Convolutional Neural Networks (CNNs) have become a common method for image fusion~\cite{chang20232,li2023learning,ma2021stdfusionnet}, such as CoCoNet~\cite{liu2024coconet}, which optimizes the fusion process through data-sensitive weights. Generative Adversarial Networks (GANs) have also demonstrated advantages in image fusion tasks~\cite{ma2019fusiongan,li2020attentionfgan,wu2023dcfusion}, like GANMcC~\cite{ma2020ganmcc}, which introduces multi-classification constraints. The Transformer model, known for handling long-range dependencies, has been applied to image fusion~\cite{ma2022swinfusion,li2025maefuse,park2023cross}, such as PromptFusion~\cite{liu2024promptfusion}, which combines Vision-Language models to enhance target recognition. Additionally, diffusion models like DDFM~\cite{zhao2023ddfm,shi2024vdmufusion,wang2025efficient} have improved the realism of fused images, and TIMFusion~\cite{liu2024task} introduces semantic segmentation to improve fusion quality under unsupervised learning. However, these methods predominantly optimize handcrafted losses and objective metrics, lacking alignment with human perceptual preferences.

\subsection{Human Feedback Reinforcement Learning} 
\noindent Reinforcement Learning from Human Feedback (RLHF) was first introduced by Christiano et al.~\cite{openai2017}, demonstrating performance that surpassed conventional deep learning paradigms. Its core lies in using human preferences as optimization objectives, training reward models to align with human value judgments, thereby guiding the learning process of policy models. RLHF was initially applied in natural language processing (NLP). InstructGPT~\cite{instructgpt}, which significantly improved GPT-3's instruction-following capability. Subsequently, models like Qwen~\cite{qwen}, DeepSeeker-V3~\cite{deepseekv3}, and LLaMA-2~\cite{touvron2023llama} have also validated the effectiveness of RLHF. Building on this, RLHF has been extended to the field of computer vision, improving tasks such as text-to-image generation~\cite{RichHuman,2024towards} and text-to-video refinement~\cite{InstructVideo}. Methods like DPOK~\cite{dpok} and ImageReward~\cite{imagereward} have enhanced the quality of generated results by providing reliable reward signals through human feedback. The success of RLHF in both NLP and CV demonstrates its effectiveness in optimizing model performance.

\section{Method}
\label{headings}

\noindent This section details our framework, including the creation of the human feedback dataset, the construction of the reward model, and the fine-tuning process of the fusion network, as illustrated in ~\cref{fig:pipeline}.

\begin{figure*}[t]
    \centering
    \includegraphics[width=1.0\linewidth]{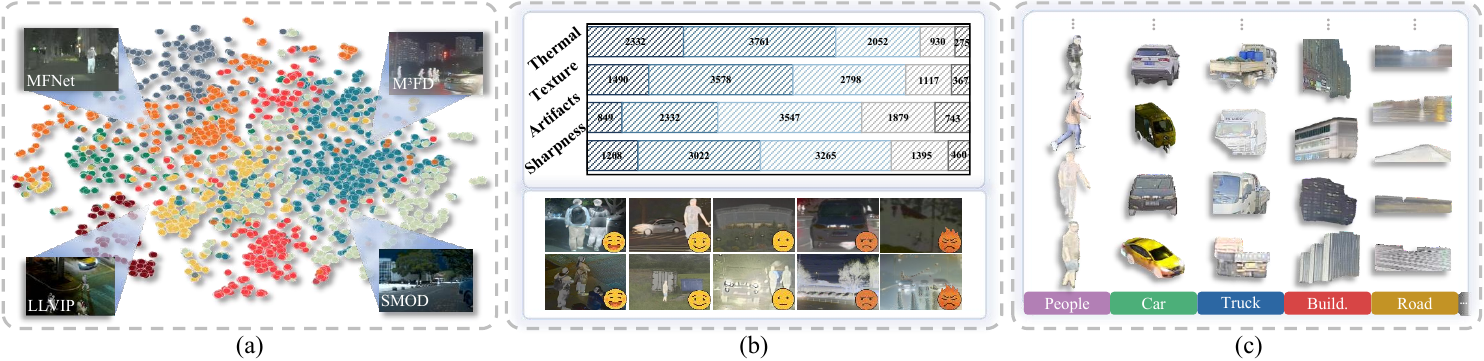}
    \caption{\textbf{Overview of our collected dataset.} (a) Data diversity: samples from multiple benchmark datasets; (b) Label diversity: each sample contains fine-grained scores across four quality dimensions and artifact heatmaps;  (c) Scene diversity: covering key semantic categories including people, cars, buildings, roads, and more.}
    \label{fig:data}
    \vspace{-3mm}
\end{figure*}

\subsection{Human Preference-Aligned Dataset for IVIF}
\label{Dataset Collection}
\noindent \textbf{Dataset Collection.}
~\cref{fig:collect} shows the process of collecting human feedback data and constructing the dataset.
To address the inconsistency between existing infrared-visible image fusion (IVIF) methods and human visual preferences in terms of visual quality, we constructed a human feedback IVIF dataset to support the modeling and evaluation of perceptual consistency. 
we first collected more than 30,000 pairs of infrared-visible images from the FMB~\cite{liu2023multi}, LLVIP~\cite{jia2021llvip}, M\(^{3}\)FD~\cite{liu2022target}, MFNet~\cite{ha2017mfnet}, RoadScene~\cite{xu2020u2fusion}, SMOD~\cite{chen2024amfd}, TNO~\cite{toet2017tno}, and VIFB~\cite{zhang2020vifb} datasets. These images cover a wide range of scenes and environmental conditions. Considering the potential presence of numerous duplicate scenes in the dataset, we used the CLIP model for data cleaning. 
This process resulted in 900 pairs of infrared-visible images. Further expert screening was conducted on the CLIP-cleaned data, ultimately yielding 850 pairs of high-quality images. we then conducted inference eleven SOTA methods: MURF~\cite{xu2023murf}, DDcGAN~\cite{xu2019learning}, MetaFusion~\cite{zhao2023metafusion}, PromptF~\cite{liu2024promptfusion}, CDDfuse~\cite{zhao2023cddfuse}, SegMif~\cite{liu2023multi}, Text-IF~\cite{yi2024text}, DDFM~\cite{zhao2023ddfm}, U2Fusion~\cite{xu2020u2fusion}, CoCoNet~\cite{liu2024coconet}, and TarDAL~\cite{liu2022target}. This resulted in the generation of 9,350 fused images.

\noindent \textbf{Large Models and Human Feedback Annotation.}
To obtain high-quality annotated fusion data that comprehensively reflects human visual preferences, we adopted a collaborative annotation process combining expert knowledge with large language models. Each fused image is assigned four fine-grained scores (1-5 scale), a total average score, and a heatmap. The fine-grained scores include: thermal retention, texture retention, artifacts level, and sharpness. Additionally, the heatmap highlights regions in the fused image where artifacts are particularly noticeable. We first invited four senior experts with over five years of IVIF research experience to perform refined scoring on 100 fused images and annotate artifact regions using center coordinates and radii. This resulted in a high-quality seed dataset of 100 images, covering the full spectrum of IVIF scenes with diverse environmental conditions and quality levels, maximizing the information density of each sample. 

GPT-4o~\cite{hurst2024gpt-4o} was trained using RLHF on large-scale multimodal datasets, learning knowledge about visual perception and human preferences, allowing the model's outputs to align with human perception. Therefore, we fine-tuned the GPT-4o model using the seed dataset to enable it to automatically assign refined scores and locate artifacts in fused images. The GPT-4o was then used to annotate all 9,350 fused images.
To ensure the accuracy of these automated annotations, after GPT-4o completed the annotations, we invited five researchers with at least three years of IVIF experience to thoroughly review and refine the GPT-generated scores and heatmaps, including correcting deviations in scoring standards and supplementing missing artifact regions. The final human feedback IVIF dataset was used to train the fusion-oriented reward model. 
More details can be found in the supplementary material.

\subsection{Fusion-Oriented Reward Model}
\label{Reward Model}

\noindent To leverage human feedback for model optimization, we train a fusion-oriented reward model on the constructed dataset. The principal aim of the reward model is to provide a comprehensive representation of both the infrared and visible images, along with their corresponding fused image, to predict four fine-grained scores and identify the spatial distribution of artifact regions in the fused image.


To fully capture the global semantic information of the input image and achieve comprehensive modeling of both structural information and semantic features, we propose a Fusion-Oriented Reward Model based on the ViT-based vision-language model. During training, the parameters of the ViT module are kept frozen, and only the upper prediction head is optimized to enhance training stability and convergence efficiency. For each input sample pair, it includes an infrared image, a visible image, and a fused image, they are respectively fed into a weight-shared ViT encoder to obtain three sets of patch-level semantic representations: $F_i=\operatorname{ViT}\left(x_i\right)[:, 1:,:]$, for $x_i \in\left\{x_{\text {ir }}, x_{\mathrm{vi}}, x_{\text {fused }}\right\}$. The above three sets of features are concatenated along the channel dimension as $\left[F_{\text {ir }}\left\|F_{\text {vi }}\right\| F_{\text {fused }}\right] \in \mathbb{R}^{N \times 3 D}$, then compressed to the original dimensionality through a linear projection and fed into another ViT encoder with the same architecture for cross-modal feature fusion. The output of the fusion ViT is reshaped into a spatial feature map format, serving as the input to the subsequent prediction branch.

The feature map $\mathcal{F}_{\text {map }} \in \mathbb{R}^{D \times H^{\prime} \times W^{\prime}}$ is used as input and fed into two separate prediction branches. For the heatmap prediction branch, the fused feature map is first channel-compressed through a convolutional module, and then progressively upsampled with residual structures to restore spatial resolution. A single-channel heatmap is finally generated using a Sigmoid activation function, producing an artifact probability map with values in the range [0,1], which highlights potential structural artifact regions in the image.
For the score prediction branch, the feature map is sequentially compressed through convolutional layers, flattened, and passed through a multilayer perceptron to regress each score value. The output is processed by a Sigmoid activation function, and the final result is a scalar representing the fine-grained score.

We train the model by jointly optimizing two loss terms. The Mean Squared Error (MSE) is used as the primary loss function. For each score dimension $s_i$ and the heatmap label $H$, the corresponding loss terms are defined as:
\begin{equation}
\mathcal{L}_{\text {score }}=\sum_{i=1}^5 \operatorname{MSE}\left(s_i, \hat{s}_i\right), \quad \mathcal{L}_{\text {heatmap }}=\operatorname{MSE}(H, \hat{H}).
\end{equation}

The final training objective is a weighted sum of the two:
\begin{equation}
\mathcal{L}_{\text {total }}=\lambda_1 \cdot \mathcal{L}_{\text {score }}+\lambda_2 \cdot \mathcal{L}_{\text {heatmap }}.
\end{equation}

\begin{figure*}[t]
    \centering
    \includegraphics[width=1.0\linewidth]{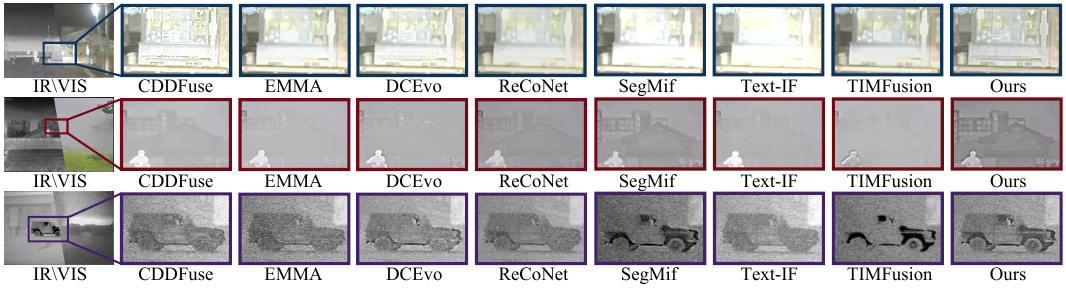}
    \caption{Qualitative comparison of our method with existing image fusion methods. From top to bottom: RoadScene, M\(^{3}\)FD, TNO.}
    \label{fusion result}
    \vspace{-3mm}
\end{figure*}

\begin{table*}[t]
  \centering
  \small
  \renewcommand\arraystretch{1}  
  \setlength{\tabcolsep}{2.2mm}      

    \caption{Quantitative comparison of IVIF between our method and state-of-the-art methods on TNO, RoadScene, and M\(^{3}\)FD datasets. \textcolor{MyRed}{\textbf{Red}} and \textcolor{MyBlue}{\textbf{blue}} represent the best and second-best results.}
    \begin{tabular}{l| cccc | cccc | cccc}
    \hline
     & \multicolumn{4}{c|}{{TNO}} & \multicolumn{4}{c|}{{RoadScene}} & \multicolumn{4}{c}{{M\(^{3}\)FD}} \\
    \cline{2-13}   \multirow{-2}{*}{{Method}}
        &CC$\uparrow$    &PSNR$\uparrow$    &Qabf$\uparrow$    &SSIM$\uparrow$           &CC$\uparrow$    &PSNR$\uparrow$    &Qabf$\uparrow$    &SSIM$\uparrow$  
        &CC$\uparrow$    &PSNR$\uparrow$    &Qabf$\uparrow$    &SSIM$\uparrow$    \\
    \hline
    CDDFuse~\cite{zhao2023cddfuse} & 0.47      & 63.49   & 0.49   & \secondbest{0.54}           & 0.52  & 59.84  & 0.43  & 0.45             & 0.62  & 63.14  & 0.61  & \best{0.51}\\ 
    DDFM~\cite{zhao2023ddfm} & 0.44         & 63.40   & 0.17   & 0.03           & 0.41  & 59.40  & 0.15  & 0.03             & \best{0.65}  & \secondbest{64.82}  & 0.48  & \secondbest{0.50}\\
    EMMA~\cite{zhao2024equivariant} & 0.45         & 63.10   & 0.42   & 0.49           & 0.50  & 59.71  & 0.42  & 0.44             & 0.58  & 63.35  & 0.57  & 0.46\\
    LRRNet~\cite{li2023lrrnet} & 0.41       & 62.38   & 0.31   & 0.41           & \secondbest{0.53}  & 59.56  & 0.39  & 0.36             & \secondbest{0.63}  & 64.18  & 0.48  & 0.38\\
    MRFS~\cite{zhang2024mrfs} & 0.49         & 61.78   & 0.41   & 0.47           & 0.50 & \secondbest{61.46}  & 0.39  & 0.45             & 0.46  & 61.24  & 0.54  & 0.44\\
    PromptF~\cite{liu2024promptfusion} & 0.46       & 62.40   & 0.51   & 0.51           & 0.52  & 61.42  & 0.49  & 0.47             & 0.50  & 62.03  & 0.60  & 0.49\\
    ReCoNet~\cite{huang2022reconet} & 0.46      & \secondbest{64.81}   & 0.34   & 0.46           & 0.52  & 60.75  & 0.36  & 0.44             & 0.58  & 64.11  & 0.47  & 0.43\\
    SegMif~\cite{liu2023multi} & \secondbest{0.50}         & 62.52   & 0.50   & 0.49           & \secondbest{0.53}  & 60.80  & \secondbest{0.51}  & 0.46             & 0.55  & 62.81  & 0.62  & 0.48\\
    SHIP~\cite{zheng2024probing} & 0.48         & 61.54   & 0.50   & 0.48           & 0.47  & 60.87  & 0.49  & 0.45             & 0.46  & 61.28  & 0.62  & 0.44\\
    TarDAL~\cite{liu2022target} & 0.47       & 63.94   & 0.43   & 0.53           & 0.48  & 60.59  & 0.40  & 0.44             & 0.57  & 63.85  & 0.41  & 0.46\\
    Text-IF~\cite{yi2024text} & 0.46      & 63.71   & 0.48   & 0.47           & 0.50  & 59.94  & 0.50  & 0.43             & 0.59  & 63.87  & \best{0.66}  & 0.48\\
    TIMFusion~\cite{liu2024task} & 0.42    & 63.98   & 0.41   & 0.49           & 0.44  & 61.12  & 0.41  & 0.44             & 0.40  & 62.17  & 0.47  & 0.38\\
    Dcevo\cite{liu2025dcevo} & 0.48        & 63.83   & \best{0.55}   & 0.53           & 0.49  & 59.66  & 0.50  & \secondbest{0.48}           & 0.55  & 62.87  & \secondbest{0.63}  & 0.47\\
    \arrayrulecolor{gray}\hdashline\arrayrulecolor{black}
    \textbf{Ours} 
    & \best{0.51} & \best{65.43} & \secondbest{0.52} &  \best{0.56} 
    & \best{0.56} & \best{61.84} & \best{0.53} & \best{0.50} 
    & \best{0.65} & \best{65.09} & \secondbest{0.63} & 0.49  \\
    \hline
    \end{tabular}%

  \label{Table: IVIF comparison}
\end{table*}

\subsection{Policy Optimization for IVIF Model}
\noindent As the foundation for our RLHF optimization, we adopt the Discriminative ross-Dimensional Evolutionary Learning (DCEvo) framework~\cite{liu2025dcevo} as our baseline fusion network. DCEvo adopts an encoder-decoder structure, using Discriminative Enhancer and Cross-Dimensional Embedding modules for feature extraction and modeling. Although existing fusion networks have achieved satisfactory performance in integrating multimodal image information, their optimization objectives are primarily based on traditional image quality metrics and task accuracy, leaving room for improvement in subjective visual quality as well as the clarity and fidelity of critical target regions.

We propose an RLHF-driven policy optimization for the IVIF model to enhance the alignment between fusion results and human perceptual preferences and improve the model's ability to perceive critical semantic targets within images. Inspired by Group Relative Policy Optimization (GRPO)~\cite{shao2024deepseekmath}, we extract key semantic regions using the Segment Anything Model (SAM)~\cite{kirillov2023segment} and compute region-level advantages. Specifically, for each pair of input visible and infrared images $(v, i)$, the fused image $F$ generated by the current fusion policy network $\pi_\theta$ is segmented by SAM into $K$ segmented images $F\left\{f_1, \ldots, f_K\right\}, f_k = F \odot \mathcal{M}_k$, where $\mathcal{M}k$ denotes the binary mask for region $k$. The segmented images, together with the fused image, are fed into the learned reward model (see Sec.~\ref{Reward Model}) for evaluation, producing a set of multidimensional quality scores $F\left\{s_1, \ldots, s_K\right\}$. Based on these scores, we compute the normalized relative advantage values within each group:
\begin{equation}
\mu = \frac{1}{K}\sum_{k=1}^K s_k, \quad 
\hat{A}_k = \frac{s_k - \mu}{\sigma + \epsilon}.
\end{equation}
Subsequently, the fusion policy network $\pi_\theta$ is updated by maximizing the following objective (loss function):
\begin{equation}
\mathcal{J}(\theta) = \mathbb{E}_{(v,i)} \left[\mathcal{L}(\theta) - \beta \cdot D_{\text{KL}}[\pi_\theta | \pi_{\text{ref}}]\right],
\end{equation}
\begin{equation}
\mathcal{L}(\theta) = \sum_{k=1}^K w_k \cdot \min\left(r_k \hat{A}_k, \text{clip}(r_k, 1-\epsilon, 1+\epsilon)\hat{A}_k\right),
\end{equation}
where $\pi_{\mathrm{ref}}$ denotes the reference policy, which is a frozen copy of the current policy $\pi_\theta$ at the initial time step. $r_k = 1 + \alpha \cdot \frac{|F_\theta[\mathcal{M}k] - F{\theta_{\text{old}}}[\mathcal{M}k]|}{|F{\theta_{\text{old}}}[\mathcal{M}k]|}$ measures the policy change for each region. $D_{\text{KL}}$ denotes the KL divergence regularization term, which prevents the policy from deviating too far from the reference policy.

\section{Experiments}

\subsection{Set up}
\begin{figure*}[t]
    \centering
    \includegraphics[width=1\linewidth]{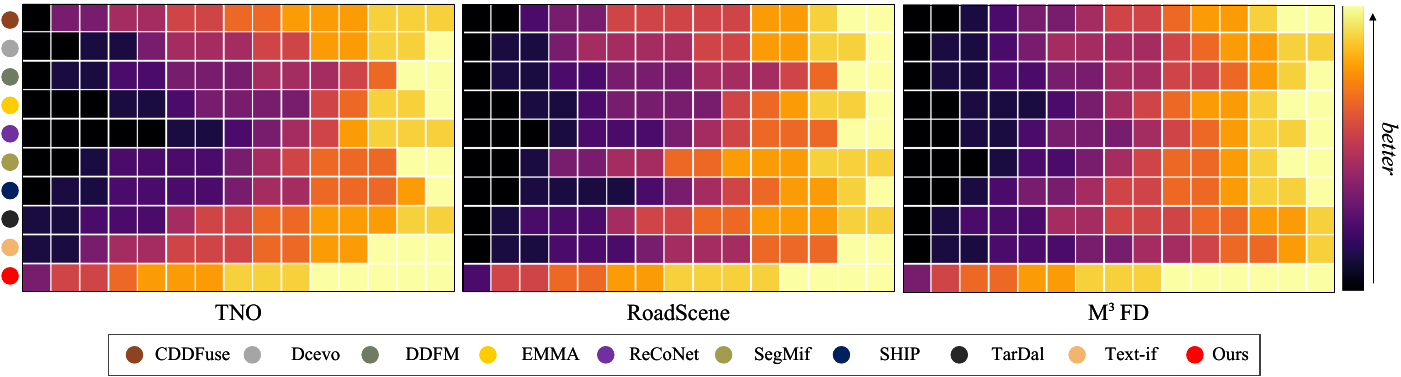}
    \caption{Preference ranking heatmaps on three datasets. Yellow represents the most preferred, and purple represents the least preferred.}
    \label{fig:heatmap}
    \vspace{-3mm}
\end{figure*}  

\begin{table}[ht]
  \centering
  \caption{Quantitative comparison of IVIF between our method and other methods using no-reference image quality assessment metrics. \textcolor{MyRed}{\textbf{Red}} and \textcolor{MyBlue}{\textbf{blue}} represent the best and second-best results.}
  \small
  \renewcommand\arraystretch{1} 
  \setlength{\tabcolsep}{0.6mm}               
    \begin{tabular}{l|cc|cc|cc}
      \hline
       & \multicolumn{2}{c|}{TNO} & \multicolumn{2}{c|}{RoadScene} & \multicolumn{2}{c}{M\(^{3}\)FD}  \\ \cline{2-7}
      \multirow{-2}{*}{{Method}} & NIQE$\downarrow$ & BRI*$\downarrow$ & NIQE$\downarrow$ & BRI*$\downarrow$ & NIQE$\downarrow$ & BRI*$\downarrow$  \\ \hline
      CDDFuse~\cite{zhao2023cddfuse} & 7.43 & 37.62 & 3.26 & 24.74 & 4.37 & 32.88  \\
      DDFM~\cite{zhao2023ddfm}       & 7.54 & 34.28 & 3.32 & 27.19 & 5.05 & 38.78  \\
      EMMA~\cite{zhao2024equivariant}& 6.33 & 26.08 & 4.93 & 22.20 & 5.45 & 35.16  \\
      LRRNet~\cite{li2023lrrnet}     & 9.31 & 38.68 & 4.42 & 29.45 & 4.33 & 37.68  \\
      MRFS~\cite{zhang2024mrfs} & 7.89         & 25.95   & 3.81   & 27.07           & 4.37 & 34.52  \\
      PromptF~\cite{liu2024promptfusion} & 6.35       & 25.06   & 3.95   & 21.93           & 4.36  & 33.09  \\
      ReCoNet~\cite{huang2022reconet}& \secondbest{5.47} & 26.18 & 5.70 & 38.86 & 5.06 & 50.22  \\
      SegMif~\cite{liu2023multi} & 6.22         & 29.12   & 3.45   & 29.41           & 4.25  & 34.75  \\
      SHIP~\cite{zheng2024probing} & 5.93         &  \secondbest{24.55}   & 3.22   & \secondbest{20.21}           & \secondbest{4.16}  & \best{27.75} \\
      TarDAL~\cite{liu2022target}   & 5.76 & 35.76 & 4.19 & 32.04  & 4.20 & 37.37  \\
      Text-IF~\cite{yi2024text}     & 6.91 & 34.50 & 3.33 & 24.82 & 5.04 & 40.06 \\
      TIMFusion~\cite{liu2024task}  & 7.19 & 32.46 & 4.44 & 30.16 & 4.61 & 46.93 \\
      DCEvo~\cite{liu2025dcevo}     & 6.25 & 33.97 & \secondbest{3.13} & 22.31 & 4.48 & 35.93 \\ 
      \arrayrulecolor{gray}\hdashline\arrayrulecolor{black}
      \textbf{Ours}                         & \best{5.37} & \best{22.58} & \best{3.06} & \best{18.79}
                                   & \best{4.03} & \secondbest{29.80}  \\ \hline
    \hline
    \end{tabular}
  \label{Table: IVIF comparison non}
  \vspace{-10pt}                             
\end{table}
\noindent \textbf{Dataset.} For the Fusion-Oriented Reward Model. As shown in ~\cref{fig:data}, our human feedback dataset contains 9,350 samples from eight  benchmark datasets, each consisting of a pair of infrared and visible images, a fused image, and corresponding fine-grained scores along with annotated artifact regions. The dataset is divided into 7,350 training samples, 1,000 validation samples, and 1,000 test samples. 
For RLHF-driven policy optimization and fusion experiments, we used the MSRS~\cite{tang2022piafusion}, RoadScene~\cite{xu2020u2fusion}, M\(^{3}\)FD~\cite{liu2022target}, and TNO~\cite{toet2017tno} datasets for training and evaluation.

\noindent \textbf{Training Setup and Evaluation.} To train the reward model, we adopt the ViT-Large-Patch16-384 model as the feature extractor, freezing its parameters, and train the score predictor and heatmap generator for 30 epochs using the AdamW optimizer with cosine annealing learning rate from 2e-5 to 1e-5 and weight decay of 2e-3. For RLHF-driven policy optimization, during fine-tuning, the KL divergence coefficient ($\beta$) is set to 0.1 and epsilon to 0.2. We use the Adam optimizer with a learning rate of 1e-4, weight decay of 0.01, batch size of 2, and train for 20 epochs. The CosineAnnealingLR scheduler is employed with a decay factor of 0.5 and a minimum learning rate of 1e-6. 

We compare our method with 13 state-of-the-art methods using four reference-based metrics: Cross-Correlation (CC), Peak Signal-to-Noise Ratio (PSNR), Quality of Fused Image Based on Brightness and Focus (Qabf), and Structural Similarity Index (SSIM)\cite{wang2004image}, along with two no-reference metrics: NIQE\cite{mittal2012making} and BRISQUE~\cite{mittal2012no}.
All experiments in the paper were conducted on two A40 GPUs.

\subsection{Infrared and visible image fusion results}
\noindent To demonstrate the effectiveness of our method, we conduct both qualitative and quantitative comparisons against 13 existing state-of-the-art methods on three benchmark datasets.

The qualitative evaluation results shown in ~\cref{fusion result} demonstrate that our method effectively preserves and enhances important features from the infrared images, such as the structure of cars and buildings, especially in low-light and foggy scenes. Additionally, guided by human feedback, our method retains more texture details, making the fused images more aligned with human visual perception. The quantitative analysis presented in ~\cref{Table: IVIF comparison} and ~\cref{Table: IVIF comparison non} shows that our method performs excellently in terms of PSNR, CC, Qabf, and SSIM, particularly achieving the highest CC and PSNR values across all three test sets. For the no-reference metrics NIQE and BRISQUE, our method also performs the best across the three test datasets, ranking first or second. These results fully demonstrate the superior fusion performance of our method.

\begin{figure*}[t]
    \centering
    \includegraphics[width=1\linewidth]{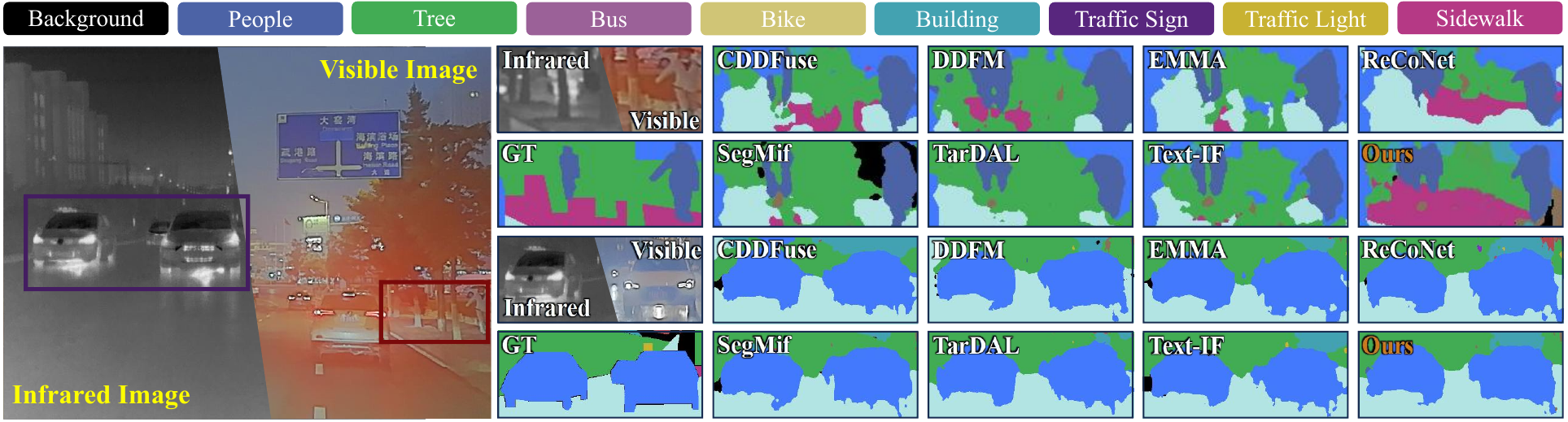}
    \caption{Qualitative comparison of our method with the fusion images generated by different fusion methods on the FMB dataset. Our approach performs the best segmentation results.}
    \label{fig:Segment}
    \vspace{-3mm}
\end{figure*}  

\begin{figure*}[t]
    \centering
    \includegraphics[width=1\linewidth]{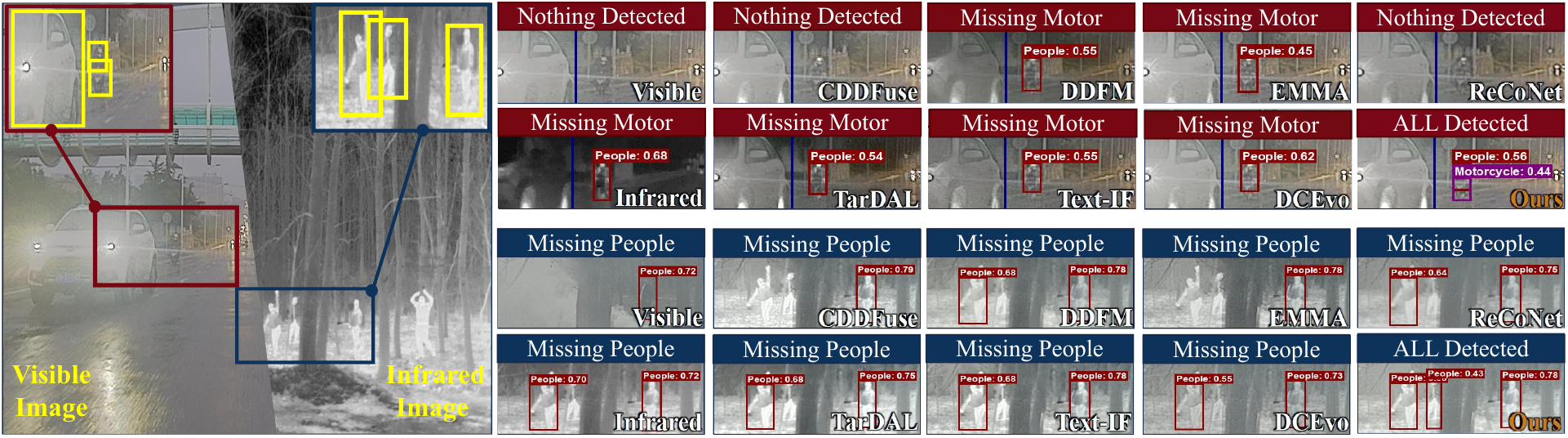}
    \caption{Qualitative comparison of our method with the fusion images generated by different fusion methods on the M\(^{3}\)FD dataset. Our method achieves the best detection results.}
    \label{fig:yolo}
    \vspace{-3mm}
\end{figure*}

\begin{table*}[ht]
  \centering
  \small
  \renewcommand\arraystretch{1}  
  \setlength{\tabcolsep}{1.6mm}   
  \caption{Quantitative comparison of our method and existing image fusion methods on downstream segmentation and detection tasks on FMB and M\(^{3}\)FD. \textcolor{MyRed}{\textbf{Red}} and \textcolor{MyBlue}{\textbf{blue}} represent the best and second-best results.}
    \begin{tabular}{l|ccccccc|ccccccc}
      \hline
      & \multicolumn{7}{c}{FMB} & \multicolumn{7}{c}{M\(^{3}\)FD} \\ \cline{2-8} \cline{9-15}
      \multirow{-2}{*}{Method} & Person & Car & Sky & Bus & Motor & Pole & mIoU & People & Car & Bus & Light & Motor & Trunk & mAP \\ \hline
      CDDFuse~\cite{zhao2023cddfuse} & 63.38 & 81.59 & 93.68 & 56.60 & 26.05 & 41.04 & 55.39 & 73.06 & 82.32 & 67.64 & 40.03 & 31.61 & 55.07 & 58.29 \\
      DDFM~\cite{zhao2023ddfm} & 63.26 & \secondbest{81.64} & \best{94.61} & 63.34 & 25.71 & \secondbest{42.60} & \secondbest{56.64} & 71.84 & 83.42 & 76.42 & \best{45.61} & 34.19 & 55.47 & 61.16 \\
      EMMA~\cite{zhao2024equivariant} & 63.10 & 80.95 & 93.85 & 62.90 & 26.96 & 40.92 & 56.00 & 72.11 & 82.32 & 75.65 & 36.24 & 33.33 & 54.11 & 58.96 \\
      LRRNet~\cite{li2023lrrnet} & 63.58 & 81.62 & 94.06 & \best{64.28} & 27.47 & 40.59 & 56.37 & 71.27 & 83.56 & 76.41 & 38.89 & \secondbest{36.15} & 54.25 & 60.09 \\
      MRFS~\cite{zhang2024mrfs} & 63.79 & 81.13 & 94.01 & 56.76 & 27.33 & 41.35 & 54.61 & 68.95 & 81.53 & 65.84 & 35.80 & 32.35 & 55.84 & 56.72 \\
      PromptF~\cite{liu2024promptfusion} & 63.80 & 80.66 & 93.91 & 55.47 & 27.88 & 40.67 & 55.42 & 73.50 & 77.47 & 76.47 & 34.82 & 32.14 & 48.72 & 56.69 \\
      ReCoNet~\cite{huang2022reconet} & 62.25 & 81.17 & 93.85 & 63.97 & 27.73 & 40.76 & 56.38 & 70.64 & 82.70 & 69.77 & 36.77 & 32.35 & 53.73 & 57.66 \\
      SegMif~\cite{liu2023multi} & 64.21 & 80.89 & 94.06 & 55.39 & \secondbest{28.65} & 42.04 & 55.99 & 73.58 & \best{84.38} & 76.47 & 40.10 & 35.46 & \secondbest{57.02} & \secondbest{61.17} \\
      SHIP~\cite{zheng2024probing} & 64.49 & 80.94 & 94.04 & 56.78 & 25.28 & 41.55 & 55.68 & 73.05 & 81.50 & 67.32 & 42.79 & 31.37 & 55.09 & 58.52 \\
      TarDAL~\cite{liu2022target} & \secondbest{64.80} & 79.73 & 94.06 & 52.99 & 26.91 & 39.97 & 54.15 & 72.90 & 81.62 & 70.26 & 35.33 & 22.35 & 51.17 & 55.61 \\
      Text-IF~\cite{yi2024text} & 64.58 & 80.77 & 93.85 & 60.56 & 28.53 & 42.43 & 56.41 & \secondbest{74.36} & 82.97 & \secondbest{76.48} & 40.93 & 34.31 & 56.36 & 60.90 \\
      TIMFusion~\cite{liu2024task} & 62.06 & 81.25 & 93.38 & 55.53 & 25.46 & 41.11 & 55.01 & 70.49 & 77.47 & 76.47 & 34.82 & 32.14 & 48.72 & 56.69 \\
      DCEvo~\cite{liu2025dcevo} & 63.51 & 80.37 & 93.65 & 52.90 & 28.35 & 41.92 & 55.88 & 73.27 & 83.20 & 66.83 & 36.78 & 33.08 & 56.24 & 58.23 \\
      \arrayrulecolor{gray}\hdashline\arrayrulecolor{black}
      Ours & \best{64.92} & \best{82.96} &  \secondbest{94.51} &  \secondbest{64.25} & \best{29.75} & \best{44.69} & \best{56.92} & \best{74.43} & \secondbest{84.17} & \best{76.58} & \secondbest{44.74} & \best{36.27} & \best{57.21} & \best{62.23} \\
      \hline
    \end{tabular}
  \vfill  
  \label{tab:combined}
  \vspace{0pt}    
\end{table*}

\subsection{Human Preference Analysis of Fusion Results} 
\noindent To further validate the effectiveness of our method in terms of human subjective preferences, we designed a subjective evaluation experiment to analyze the fusion quality of different IVIF methods. We selected 15 pairs of representative images from 10 datasets and invited 15 independent participants (including 5 domain experts and 10 non-experts) to perform a double-blind evaluation. The evaluators were asked to rank their preferences for each pair of fused images (10 images in total). The final results were obtained by normalizing the preference rankings of all evaluators and calculating the average ranking for each fused image pair.

~\cref{fig:heatmap} shows the preference ranking heatmap for different methods on three datasets. The color intensity in the heatmap visually reflects the performance of each method, with yellow indicating the best and purple indicating the worst. 
From the heatmap, it is evident that our method achieves higher average rankings across all datasets, demonstrating that the fused images generated by our method better align with human preferences.

\subsection{Downstream IVIF Applications} 
\noindent We further explored the performance of the fused images in downstream tasks, using the FMB and M\(^{3}\)FD datasets for semantic segmentation and object detection.

~\cref{fig:Segment} and~\cref{fig:yolo} show the qualitative evaluation results for segmentation and detection tasks. For semantic segmentation, our method demonstrates higher classification accuracy in both low-light and high-light scenes. For object detection, our method successfully detects the motorbike in low-light conditions and accurately detects people in the fog under dense fog conditions, while other methods missed some targets. ~\cref{tab:combined} presents the quantitative results for both tasks. For semantic segmentation, our method outperforms other methods across multiple classification tasks. For key targets such as car and person, our segmentation accuracy ranks first. 
For object detection, our method achieves the highest mAP. It ranks first in four detection categories, indicating its ability to provide richer feature information and improve detection accuracy. 

\subsection{Ablation studies}
\begin{figure}[t]
    \centering
    \includegraphics[width=\linewidth]{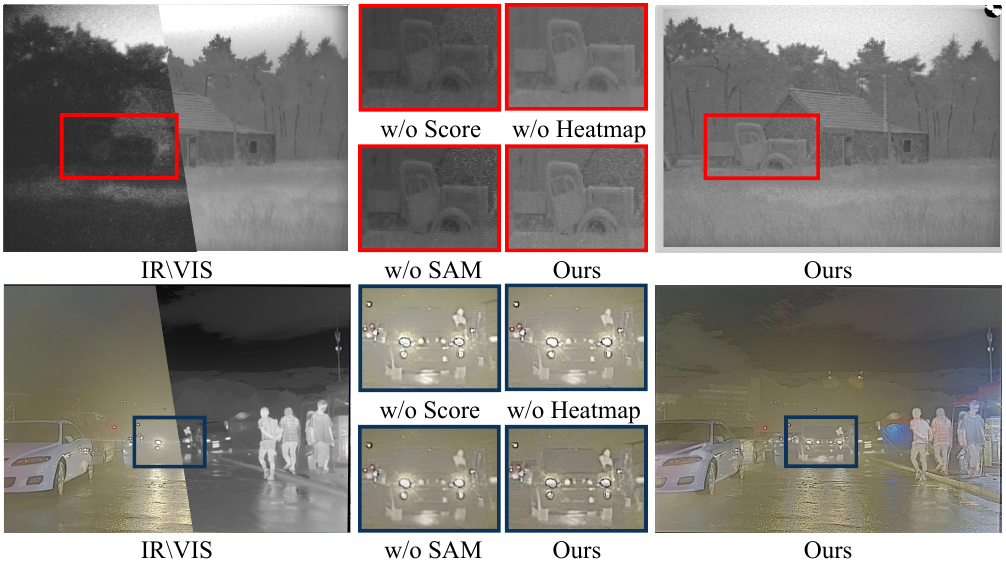}
    \caption{Comparison between w/o score, w/o heatmap, w/o SAM, and our method.}
    \label{fig:ablation2}
\end{figure}

\noindent \textbf{Key Components Ablation.} To validate the effectiveness of each component in our proposed framework, we conducted comprehensive ablation studies. First, for the fusion-oriented reward model, we removed the score prediction branch and the heatmap prediction branch separately. Secondly, during the fine-tuning process, we removed the SAM segmentation and replaced the semantic regions with randomly cropped source image patches.
~\cref{fig:ablation2} shows the results of the ablation experiments. Our method preserves clear details under low-light conditions. Removing the score prediction and heatmap prediction branches leads to blurred vehicle edges and artifacts, as the reward model cannot fully evaluate the fused image quality. Removing SAM segmentation lacks weighted optimization for important semantic regions, resulting in worse performance compared to our method.
~\cref{tab:ablation-comparison2} qualitatively shows the results of the ablation experiments on three datasets. Our method consistently demonstrated the best performance on all three datasets, and the removal of any module led to a decline in the corresponding metrics. This validates the effectiveness of each module in our method. 

\begin{table}[t]
    \centering
    \small
    \renewcommand\arraystretch{1.0}
    \setlength{\tabcolsep}{1.4mm}
    \caption{Comparison of ablation studies removed key components.}
    \begin{tabular}{l|cc|cc|cc}
        \hline
         & \multicolumn{2}{c|}{TNO} & \multicolumn{2}{c|}{RoadScene} & \multicolumn{2}{c}{M$^{\scalebox{0.7}{3}}$FD} \\
         \cline{2-7}
        \multirow{-2}{*}{{Method}} & CC$\uparrow$ & PSNR$\uparrow$ & CC$\uparrow$ & PSNR$\uparrow$ & CC$\uparrow$ & PSNR$\uparrow$ \\
        \hline
        w/o Score& 0.50	 & 64.21 & 0.51 & 60.97 & 0.58 & 64.81 \\
       w/o Heatmap     & 0.50 & 65.17 & 0.54 & 61.21 & 0.60 & 64.93 \\
       w/o SAM      & 0.48 & 65.03 & 0.52 & 60.92 & 0.57 & 63.02 \\
        \arrayrulecolor{gray}\hdashline\arrayrulecolor{black}
        \rowcolor[gray]{0.9} \textbf{Ours} 
                 & \textbf{0.51} & \textbf{65.43}
                 & \textbf{0.56} & \textbf{61.84}
                 & \textbf{0.65} & \textbf{65.09} \\
        \hline
    \end{tabular}
    \label{tab:ablation-comparison2}
    \vspace{-10pt}  
\end{table}

\begin{figure}[t]
    \centering
    \includegraphics[width=\linewidth]{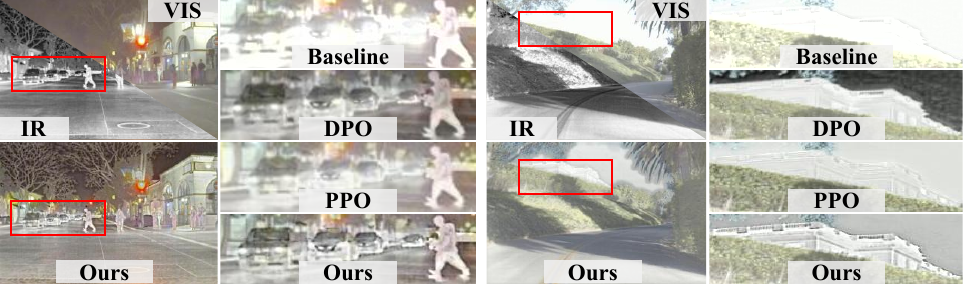}
    \caption{Comparison between the baseline, DPO~\cite{rafailov2023direct}, PPO~\cite{schulman2017proximal}, and our method.}
    \label{fig:ablation}
\end{figure}

\noindent \textbf{Policy Optimization Strategy Ablation.} We conducted ablation experiments to validate the effectiveness of our proposed fusion method based on GRPO fine-tuning and human feedback guidance. For comparison, we compare the Direct Preference Optimization (DPO)~\cite{rafailov2023direct} and Proximal Policy Optimization (PPO)~\cite{schulman2017proximal} optimization baseline methods. Qualitative and quantitative analysis was performed on three datasets.
As shown in ~\cref{fig:ablation}, our method retains the building contour information in the infrared image as well as the texture and color information in the visible image. It balances the infrared and visible information, generating images that are more aligned with human visual preferences. The qualitative evaluation results in ~\cref{tab:ablation-comparison} demonstrate the performance on the three datasets. Compared to the baseline methods, although DPO and PPO methods show improvements, our method performs the best, validating its effectiveness.

\begin{table}[t]
    \centering
    \small
    \renewcommand\arraystretch{1.0}
    \setlength{\tabcolsep}{1.8mm}
    \caption{Comparison of ablation studies with different fine-tuning methods.}
    \begin{tabular}{l|cc|cc|cc}
        \hline
         & \multicolumn{2}{c|}{TNO} & \multicolumn{2}{c|}{RoadScene} & \multicolumn{2}{c}{M$^{\scalebox{0.7}{3}}$FD} \\
         \cline{2-7}
        \multirow{-2}{*}{{Method}} & CC$\uparrow$ & PSNR$\uparrow$ & CC$\uparrow$ & PSNR$\uparrow$ & CC$\uparrow$ & PSNR$\uparrow$ \\
        \hline
        Baseline & 0.48 & 63.83 & 0.49 & 59.66 & 0.55 & 62.87 \\
        DPO      & 0.50 & 63.98 & 0.49 & 61.50 & 0.57 & 63.74 \\
        PPO      & 0.51 & 63.59 & 0.53 & 61.17 & 0.59 & 64.32 \\
        \arrayrulecolor{gray}\hdashline\arrayrulecolor{black}
        \rowcolor[gray]{0.9} \textbf{Ours} 
                 & \textbf{0.51} & \textbf{65.43}
                 & \textbf{0.56} & \textbf{61.84}
                 & \textbf{0.65} & \textbf{65.09} \\
        \hline
    \end{tabular}
    \label{tab:ablation-comparison}
    \vspace{-10pt}  
\end{table}


\section{Conclusion}

\noindent We propose a fusion method based on RLHF that aligns image fusion quality with human visual preferences through human feedback. On the one hand, we design a human feedback dataset that includes multidimensional scores and image artifact annotations, and train a Fusion-Oriented reward model, filling the gap in existing IVIF research regarding datasets and metrics related to human preferences. On the other hand, by combining the trained reward model and inspired by GRPO, we fine-tune the IVIF task, overcoming the limitations of traditional fusion networks in capturing human visual preferences. In the end, the generated fused images are more aligned with human visual preferences, significantly improving the quality of the fused images.

{
    \small
    \bibliographystyle{ieeenat_fullname}
    \bibliography{main}
}


\end{document}